\documentclass[lettersize,journal]{IEEEtran}
\usepackage{amsmath,amsfonts}
\usepackage{algorithmic}
\usepackage{algorithm}
\usepackage{array}
\usepackage{textcomp}
\usepackage{stfloats}
\usepackage{url}
\usepackage{verbatim}
\usepackage{graphicx}
\usepackage{cite}
\usepackage{color}
\usepackage{booktabs}
\usepackage{xcolor}
\usepackage{multirow}
\usepackage{enumitem}
\usepackage{color}
\usepackage{xcolor}
\definecolor{citecolor}{HTML}{0071bc} 
\usepackage[colorlinks, linkcolor=red,  anchorcolor=blue, citecolor=citecolor]{hyperref}
\usepackage{float}
\usepackage{xcolor}
\definecolor{SeaGreen4}{RGB}{0,205,102} 
\definecolor{SlateBlue}{RGB}{106,90,205} 
\definecolor{DarkRed}{RGB}{178,34,34} 
\usepackage{times}
\usepackage{soul}
\usepackage{url}
\usepackage[utf8]{inputenc}
\usepackage[small]{caption}
\usepackage{graphicx}
\usepackage{amsmath}
\usepackage{amsthm}
\usepackage{booktabs}
\usepackage{algorithm}
\usepackage{algorithmic}
\usepackage[switch]{lineno}
\usepackage{color}
\usepackage{xcolor}
\usepackage{textcomp}
\usepackage{booktabs} 
\usepackage{amssymb}
\usepackage{pifont}
\newcommand{\cmark}{\ding{51}}%
\newcommand{\xmark}{\ding{55}}%
\usepackage{makecell}
\usepackage{colortbl}
\definecolor{mygray}{gray}{.9}
\definecolor{mypink}{rgb}{.99,.91,.95}
\definecolor{mycyan}{cmyk}{.3,0,0,0}

\begin{document}

\title{VcT: Visual change Transformer for Remote Sensing Image Change Detection}


\author{Bo Jiang, Zitian Wang, Xixi Wang, Ziyan Zhang, Lan Chen, Xiao Wang*,~\emph{Member,~IEEE}, Bin Luo*,~\emph{Senior Member,~IEEE}

\thanks{Bo Jiang is with the Information Materials and Intelligent Sensing Laboratory of Anhui Province, School of Computer Science and Technology, Anhui University, Hefei 230601, China (jiangbo@ahu.edu.cn)}
\thanks{Zitian Wang, Xixi Wang, Ziyan Zhang, Xiao Wang, and Bin Luo are with School of Computer Science and Technology, Anhui University, Hefei 230601, China. (email: xiaowang@ahu.edu.cn, luobin@ahu.edu.cn)} 
\thanks{Lan Chen is with School of Electronic and Information Engineering, Anhui University, Hefei 230601, China. (email: chenlan@ahu.edu.cn)} 
\thanks{* denotes Corresponding Author}}



 \markboth{IEEE Transactions on Geoscience and Remote Sensing 2023}%
 {Shell \MakeLowercase{\textit{et al.}}: A Sample Article Using IEEEtran.cls for IEEE Journals}


\maketitle

\begin{abstract}
Given two remote sensing images, the goal of visual change detection task is to detect significantly changed areas between them. Existing visual change detectors usually adopt CNNs or Transformers for feature representation learning and focus on learning effective representation for the changed regions between images. Although good performance can be obtained by enhancing the features of the change regions, however, these works are still limited mainly due to the ignorance of mining the unchanged background context information. It is known that one main challenge for change detection is how to obtain the consistent representations for two images involving different variations, such as spatial variation, sunlight intensity, etc. In this work, we demonstrate that carefully mining the common background information provides an important cue to learn the consistent representations for the two images which thus obviously facilitates the visual change detection problem. Based on this observation, we propose a novel Visual change Transformer (VcT) model for visual change detection problem. To be specific, a shared backbone network is first used to extract the feature maps for the given image pair. Then, each pixel of feature map is regarded as a graph node and the graph neural network is proposed to model the structured information for coarse change map prediction. Top-K reliable tokens can be mined from the map and refined by using the clustering algorithm. Then, these reliable tokens are enhanced by first utilizing self/cross-attention schemes and then interacting with original features via an anchor-primary attention learning module. Finally, the prediction head is proposed to get a more accurate change map. Extensive experiments on multiple benchmark datasets validated the effectiveness of our proposed VcT model. The source code and pre-trained models is available at \url{https://github.com/Event-AHU/VcT_Remote_Sensing_Change_Detection}. 
\end{abstract}


\section{Introduction}
\IEEEPARstart{R}{emote} sensing image change detection targets finding the variable pixel-level regions between given two images, such as optical, multispectral, infrared, and synthetic aperture radar (SAR) images captured at long intervals~\cite{singh1989review}. It is one of the most important research topics in the pattern recognition and computer vision communities and has been widely used in many applications~\cite{ban2012multitemporal, kennedy2009remote, hou2017change, gupta2019creating}. 
Although significant developments have been achieved, remote sensing change detection is still a challenging and difficult task due to the following two issues. The first one is that different remote sensing systems have different temporal, spatial, spectral, and radiometric resolutions which make 
the comparison and analysis between different images be very challenge. The second one is the environmental factors, such as sunlight intensity, atmospheric and soil moisture, which will lead to image  degradation. Influenced by these issues, the same object may show different spectral characteristics. 
Recently, with advancements in technology and application demands, satellite sensors have witnessed significant improvements in their capabilities. This progress has allowed us to acquire a larger quantity of very high-resolution optical remote sensing images.
Consequently, optical remote sensing images have emerged as the preferred data source for change detection problem.

\begin{figure}
\centering
\includegraphics[width=0.48\textwidth]{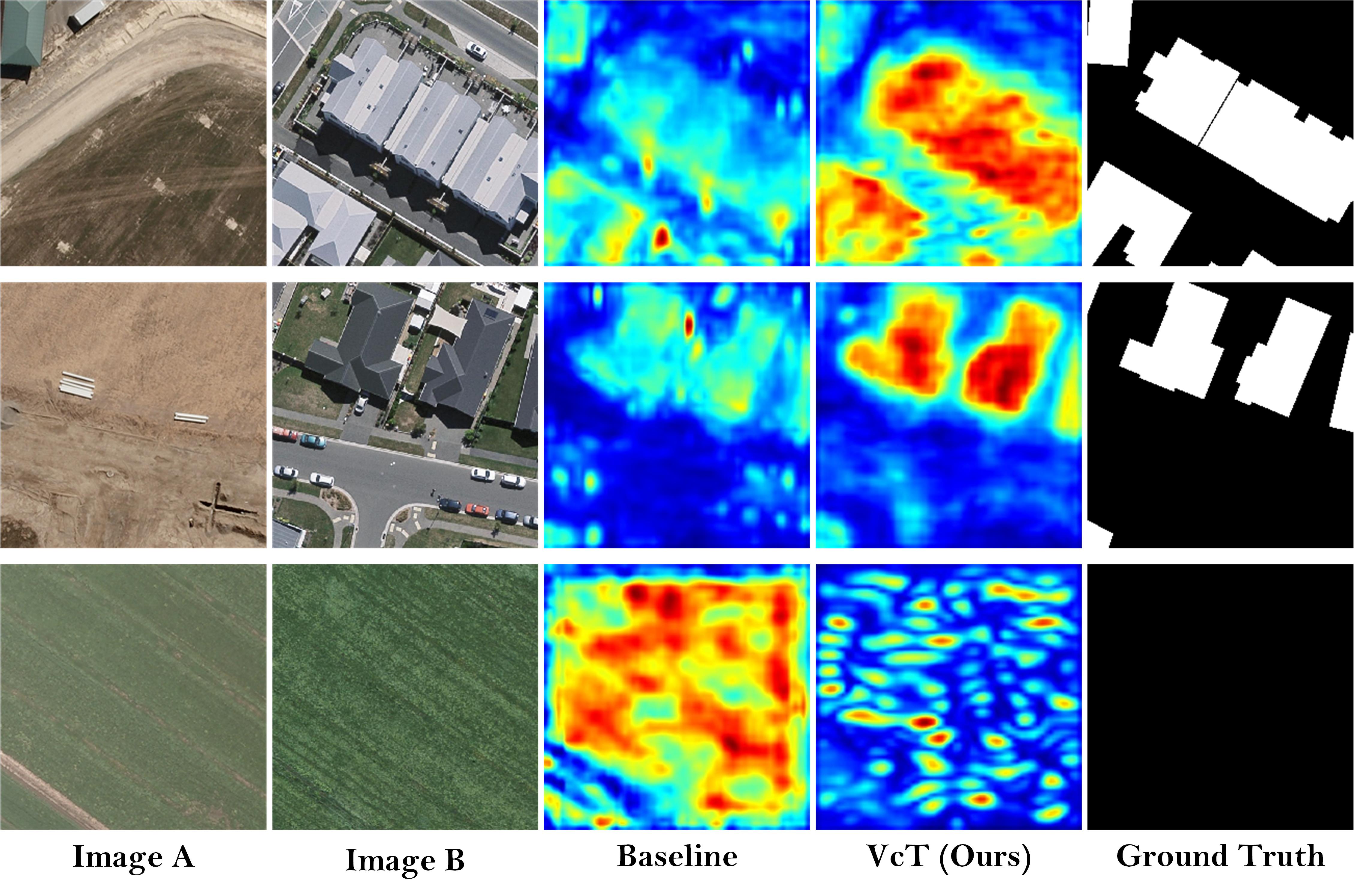} 
\caption{We compare the baseline remote sensing image change detector with our proposed VcT. The visualized feature map corresponds to the probability map generated by the final prediction head output.}  
\label{Fig.1}
\end{figure}

More and more researchers are devoted to this research problem and many convolutional neural network (CNN) based models are proposed~\cite{chen2020spatial, park2022dual, chen2020dasnet, zhang2018triplet, zhang2020feature}. Subsequently, many schemes are proposed to further improve the reception field (RF) of convolution layers, including convolutional layers stacking~\cite{chen2020spatial, park2022dual}, dilated convolution~\cite{zhang2018triplet}, and attention mechanisms~\cite{chen2020spatial, chen2020dasnet}. The essence of the attention mechanism is to assign more weights to the information of interest which thus can suppress the useless background information.
In detail, existing  models can be categorized as  three types, i.e., the spatial attention based method~\cite{liu2020building, zhang2020deeply}, channel attention based method~\cite{liu2020building, zhang2020deeply}, and self-attention based models~\cite{chen2020spatial, diakogiannis2021looking}. However, existing works generally either employ attention learning on each pair of images separately or simply use attention mechanism in the spatial/channel dimension to fuse the dual temporal modalities. Therefore, their performances are still limited mainly due to the usage of the local convolutional filters.

Recently, self-attention and Transformers have drawn more and more attention due to their strong ability on global range feature learning and modeling. Both natural language processing and computer vision tasks are dominated by this approach. There are also some recent algorithms to exploit Transformer for remote sensing change detection problem~\cite{chen2021remote, bandara2022Transformer, zhang2022swinsunet}. To be specific, Chen et al.~\cite{chen2021remote} represent the CNN features as semantic tokens and attempt to learn the context information using Transformer encoder. Then, the learned features are embedded into pixel space through Transformer decoder network. Bandara et al.~\cite{bandara2022Transformer} propose a Siamese network architecture that contains a hierarchical Transformer encoder and MLP (Multi-Layer Perceptron) decoder, which achieves good performance without a CNN backbone network. Zhang et al.~\cite{zhang2022swinsunet} design a complete Transformer network for remote sensing change detection based on the Swin Transformer network. 
Despite achieving better performance than CNN-based models, we think current issues still exist in the aforementioned models.
In general, existing works mainly focus on enhancing the representation of the changed regions in two images but ignore the unchanged background areas. Therefore, the influence of unconspicuous changes in the unchanged regions may be magnified which may cause the detector to judge unchanged areas with relatively large differences as changed areas. 
It is known that one main challenge for visual change detection problem is how
to obtain the consistent representations for the input two images. 
This inspires us to think about how to take advantage of the unchanged information to suppress the irrelevant cluttered changes and make the final results more reliable. 

In this paper, we demonstrate that mining unchanged background tokens
provides an important cue to learn the consistent representations
for the two images which thus obviously facilitates the visual
change detection problem. 
Based on this observation, we propose a novel Visual change Transformer (VcT) framework for
the visual change detection problem. 
To be specific, we first extract their feature maps by using a shared backbone network (the modified ResNet18~\cite{he2016deep} is adopted). Then,  each pixel of feature map is regarded as a
graph node and the Graph Neural Network (GNN) is employed to model
the structured information for the coarse change map prediction. 
After that, top-K reliable tokens can be selected from the coarse map and refined
by using the k-means clustering algorithm~\cite{likas2003global}.  
 To further enhance the local and global relations, we propose the self-attention operation to encode the clustered features  and split them into dual groups for corresponding images respectively for feature interactive learning by using the cross-attention module. Finally, a new anchor-primary attention module is introduced to achieve enhancement between the newly generated tokens and backbone features. The decoder module is utilized to output the final change map. According to the probability maps visualized in Fig.~\ref{Fig.1}, we can find that the undesired effect of irrelevant changes can be well reduced by our newly proposed VcT model. 

To sum up, the contributions of this paper can be summarized as the following three main aspects:

$\bullet$ In contrast to previous methods that commonly employ the Visual Transformer (ViT) as the backbone for extracting feature representations, we introduce a new remote sensing change detection framework called Visual change Transformer (VcT). This framework effectively utilizes both intra-image and inter-image cues by capturing the dependencies between reliable tokens in dual images. 


$\bullet$ We introduce a novel module for token selection called Reliable Token Mining (RTM), which utilizes a graph neural network (GNN) to consistently identify reliable background tokens from dual images. Unlike previous Transformer-based visual change detection approaches that rely on manually established tokens, our method automates the selection process through our designed RTM, enhancing the efficiency and accuracy of the detection process.


$\bullet$ Extensive experiments on multiple widely used remote sensing change detection benchmarks validate the effectiveness of our proposed VcT model.

The organization of this paper is described as follows. In Section~\ref{relatedworks}, we give an introduction to the related works on remote sensing image change detection, and Transformer networks. Then, we introduce our newly proposed reliable token mining based Transformer framework for remote sensing image change detection in Section~\ref{methods}. After that, we conduct extensive experiments to validate the effectiveness of our proposed modules in Section~\ref{experiments}. Finally, we analyze the limitations of our model and provide some possible future works in Section~\ref{limitAnalysis} and conclude this paper in Section~\ref{conclusion} respectively.

\section{Related Work} \label{relatedworks} 

In this section, we provide a brief introduction to Remote Sensing Image Change Detection and Transformer Networks. For further information on these two aspects, one can refer to the  survey papers~\cite{han2022Formersurvey, shafique2022RSCDSurvey}.  

\textbf{Remote Sensing Image Change Detection. } 
Remote Sensing Image Change Detection can be divided into two categories, i.e., traditional-based methods and deep learning-based approaches. 
The  traditional methods have been designed which include algebraic algorithms~\cite{li2016change,lambin1994change}, classification method~\cite{peiman2011pre} and transformation methods~\cite{nielsen1998multivariate,celik2009unsupervised}. 
The main disadvantage of this approach is that it is not robust enough and also generally depends on the accuracy of the classification. 
Lu et al. proposed a method called Change Detection with Markov Random Field (CDMRF) for change detection \cite{lu2019landslide}. The method combines normalized vegetation index, principal component analysis, independent component analysis, and Markov random field  together for the landslide change detection. Pu et al. conducted change detection of invasive species using direct multi-temporal image classification \cite{pu2008invasive}. They compared the performance of two classifiers, artificial neural networks and LDA, and found that artificial neural networks outperformed LDA \cite{pu2008invasive}. Traditional change detection methods mainly rely on manual feature extraction \cite{nielsen2007regularized, li2015gabor, gong2011change, moser2006generalized}. These methods are usually highly interpretable, but they generally depend on manual feature extraction. 



Existing state-of-the-art change detectors for remote sensing images are developed on the basis of deep neural networks. 
The first type of detectors follows a two-stage based approach, where the images are first classified and then compared to obtain the final changed results~\cite{ji2019building, nemoto2017building, liu2019temporal}. However, this approach has a drawback as it necessitates obtaining additional classification tags and semantic labels, which can be expensive. For example, certain researchers~\cite{ji2019building, liu2019temporal} initially segment each image independently to acquire the semantic labels, and subsequently consider inconsistent labels for the same regions as changed regions. While these approaches seem intuitive, the need for semantic labels escalates the cost of data annotation.

The second solution involves single-stage based methods, which are more efficient and capable of directly generating change results by integrating bitemporal information. These single-stage models~\cite{daudt2018fully, wu2021multiscale} can predict the changed regions directly by fusing bitemporal information, resulting in higher efficiency. The patch-level algorithms formulate the change detection as a similarity detection problem by chunking the bitemporal images into many patches and then getting the central predictions. Daudt et al. exploit the application of convolutional neural networks for urban change detection to classify each patch~\cite{daudt2018urban}. Rahman et al. present a patch-based Siamese neural network, aiming to detect structural changes in objects~\cite{rahman2018siamese}. Wang et al. propose a method based on the deep Siamese convolutional network to explore the effect of patch size on detection accuracy~\cite{wang2020deep}.

Compared to the patch-level approach, pixel-level change detection algorithms are more effective and can directly generate a pixel-level change map. To be specific, Fang et al. proposed a dual learning-based Siamese framework (DLSF), which highlights the pixel-level differences in the change region and then focuses on detecting the change region~\cite{fang2019dual}. Daudt et al. propose two Siamese extensions of fully convolutional networks which is able to learn pixel-level changes from scratch~\cite{daudt2018fully} for change detection. In addition to the aforementioned CNN-based models, there are also some works developed based on Generative Adversarial Networks (GAN) and Graph Convolutional Networks (GCN). For example, Liu et al. propose a supervised domain adaptation framework called SDACD for cross-domain change detection, which uses GAN to perform cross-domain style transformation of images, thus effectively narrowing the domain gap in a generative manner with circular consistency constraints~\cite{liu2022end}. Noh et al. propose image reconstruction loss, using only an unlabeled single image as training input and generating another by GAN. The network uses reconstruction loss values as a detection criterion~\cite{noh2022unsupervised}. Ali et al. propose a novel graph formulation (BLDNet) and use GCN learning relationships and representations from both non-stationary neighborhoods and local patterns~\cite{ismail2022bldnet}. There are also works built based on attention schemes which will be introduced in the next subsection.


 \textbf{Transformer Networks. }
The key component of the Transformer network is the self-attention mechanism which models the long-range relations of the input tokens well~\cite{vaswani2017attention}. It is firstly proposed to handle the translation tasks in the natural language processing community and achieves significant improvements compared with widely used recurrent neural network (RNN) based models. 
Inspired by the great success of self-attention and Transformer, some researchers also attempt to migrate this model for computer vision tasks. For example, Lee et al. propose the Set Transformer which designs a novel attention mechanism to model interactions among elements for the input set~\cite{lee2019set}. Jiang et al. propose a novel efficient Anchor Matching Transformer (AMatFormer) which conducts self-/cross-attention on some anchor features and leverages these as message bottleneck to learn the representations for all primal features~\cite{jiang2023amatformer}. 
Many representative Transformer models are proposed for backbone feature extraction (such as ViT~\cite{dosovitskiy2020image}, Swin Transformer~\cite{liu2021swin}), and are widely used in many downstream tasks, like segmentation~\cite{liu2021swin, liu2022swin}, detection and tracking~\cite{sun2020transtrack, xu2021transcenter, wang2022BeamTrack, tang2022coesot}, and generation domain~\cite{ding2021cogview, kitaev2020reformer}. There are also many researchers who adopt Transformer networks for multi-modal feature learning (such as RGB, language, audio, and event stream)~\cite{wang2023mmptms, wang2021mutualformer}. These works fully demonstrate the effectiveness and generalization of Transformers for various data inputs.

There are also some researchers who exploit the Transformer models for visual change detection tasks~\cite{liu2020building, zhang2020deeply, jiang2020pga, chen2021remote, bandara2022Transformer, xu2021vitae, cheng2022isnet, wang2022empirical,10139838,10137749,ghaderi2022siamixformer}. The introduction of attention mechanisms for contextual modeling is essential for identifying changes, and the learning of global relational information can better enhance features. For example, Liu et al. construct dual attention modules (DAM) to improve feature representation using spatial and channel dependencies~\cite{liu2020building} and Zhang et al. propose a network in which multi-level depth features of the original image are fused with image difference features through an attention module~\cite{zhang2020deeply}. 
Jiang et al. propose an attention-guided Siamese network based on pyramidal features~\cite{jiang2020pga}. Cheng et al. propose a deep network with improved separability (ISNet), which refines features by employing the strategies of margin maximization and attention mechanisms~\cite{cheng2022isnet}. Chen et al. extract semantically-tagged visual words and use the Transformer network to model the context in spacetime and enhance the region of interest~\cite{chen2021remote}. Bandara et al. present a Siamese network consisting of Transformer blocks and the network efficiently provides the multiscale features needed for accurate change detection through a hierarchical structure. In addition, a simple Multi-Layer Perceptron (MLP) decoder was constructed~\cite{bandara2022Transformer}. Wang et al. pre-train the improved ViTAE model~\cite{xu2021vitae} with a remote sensing dataset and demonstrate good performance on the detection task~\cite{wang2022empirical}. 
Zhang et al. introduce a novel attention mechanism called Cross-Temporal Difference (CTD), which analyzes relation changes in multi-temporal images. They also design Consistency-Perception Blocks (CPBs) to generate the desired change map~\cite{10139838}. Fu et al. propose a Differential Feature Extraction Network based on Adaptive Frequency Transformer (AFFormer). This network separates change targets and environments from a frequency perspective, providing richer and more detailed information for remote sensing change detection tasks~\cite{10137749}. 
Ghaderi et al. propose a Transformer Siamese  network as well, termed SiamixFormer, for building detection~\cite{ghaderi2022siamixformer}. 

Different from previous related works, our proposed VcT framework considers the invariant background information and introduces a novel reliable token mining (RTM)  module for reliable token selection. 
Based on RTM, we develop a novel Transformer architecture to carefully model the relationships of the selected tokens representing the unchanged regions instead of enhancing change regions focused in many previous related works.

\section{Our Proposed Approach} \label{methods} 
In this section, we will first give an overview of our proposed Visual change Transformer (VcT) framework for remote sensing image. Then, we will dive into the details of our proposed framework with a focus on the input embedding, reliable token mining, self-/cross-attention interaction module, anchor-primary attention, prediction head, and loss function.

\begin{figure*} 
\centering
\includegraphics[width=1\textwidth]{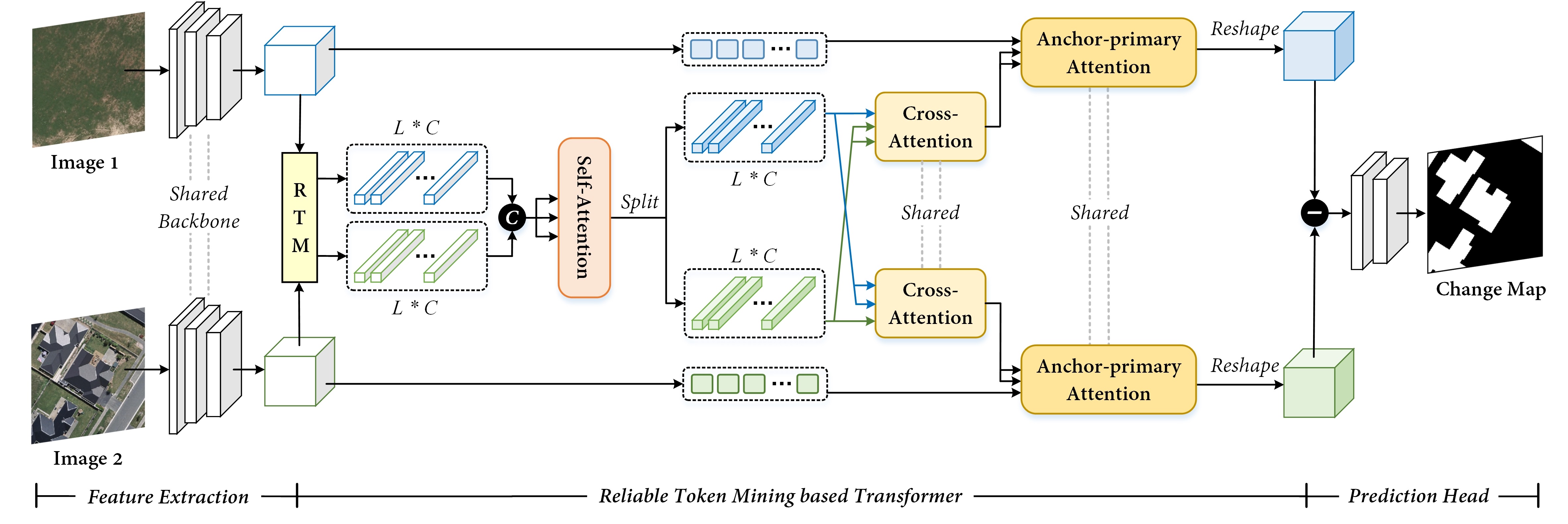} 
\caption{\textbf{An overview of our proposed Visual change Transformer (VcT) for remote sensing image change detection.} It mainly contains four modules, i.e., the shared backbone network, reliable token mining module, self-/cross-attention feature enhancement module, and CNN decoder. Given the input images, we first adopt a shared ResNet18 as the backbone network for feature embedding. Then, a novel Reliable Token Mining (RTM) module is proposed to mine the tokens of length $L$ derived from the clustering algorithm for change detection purpose. Then, self-attention and cross-attention are used for intra-relation mining and inter-relation feature learning, respectively. We adopt another anchor-primary attention scheme to fuse the selected features and original backbone features. After that, the dual enhanced features are subtracted and transformed into the change map using a CNN decoder network. }
\label{framework}
\end{figure*}

\subsection{Overview}
As illustrated in Fig.~\ref{framework}, our proposed VcT framework consists of four main modules: the backbone network, reliable token mining module, self-/cross-/anchor-primary attention module, and CNN decoder network. Given the input of two  images, we extract the feature descriptors by using a shared backbone network. The modified ResNet18~\cite{he2016deep} is adopted in our experiments. Next, we feed the features into the Reliable Token Mining (RTM) module to obtain tokens of length $L$ by using a clustering algorithm for change detection. The output features are then concatenated and fed into the self-attention module for intra-relation mining. Cross-attention layers are utilized to achieve inter-relation feature learning. An anchor-primary attention module is adopted to fuse the selected features and original backbone features. Finally, we apply a subtract operation on the dual enhanced features and output the change map using a CNN decoder network.

\subsection{Network Architecture} 

In this subsection, we introduce the main parts of our network, i.e., Input Embedding Module, Reliable Token Mining Module, Self-Attention Module, Cross-Attention Module, Anchor-Primary Attention, and Prediction Head.

\noindent 
\textbf{Input Embedding.}
Given the dual input images $I_1 \in \mathbb{R}^{H_{0}\times W_{0}\times 3}$ and $I_2 \in \mathbb{R}^{H_{0}\times W_{0}\times 3}$ for change detection, where $H_0$ and $W_0$ denote the height and width of input images respectively, we adopt the ResNet18~\cite{he2016deep} as the shared backbone network with slight modification for feature embedding. The output feature maps are denoted as $X_{1} \in \mathbb{R}^{H\times W\times C}, X_{2} \in \mathbb{R}^{H\times W\times C}$, where $C$ is the number of channels of feature maps. As we know, the used CNN backbone only learns the local feature, and existing works transform the features maps into tokens and demonstrate that the self-attention based Transformer captures the global features well. However, we believe that not all tokens are desired for the final change detection results. 
To address this issue, we propose the Reliable Token Mining (RTM) module to select reliable tokens, as introduced below.

\noindent 
\textbf{Reliable Token Mining (RTM). }
For visual change detection tasks, it is desired to select some reliable tokens (ideally from unchanged regions) to achieve the information communication between two images. To achieve this purpose, we need to understand which regions are unchanged. Thus, we attempt to obtain a detector independent of the prediction head to get a coarse change map as the prior knowledge. In our implementation, we propose to employ the graph convolution network (GCN)~\cite{kipf2016semi} module and utilize K-means clustering to select some valid tokens effectively. The K-means algorithm is simple and effective and will not increase the number of parameters in the model. According to our experimental results, the coarse map obtained using GCN is closer to the final change map with higher accuracy. 

\begin{figure}
\centering
\includegraphics[width=0.48\textwidth]{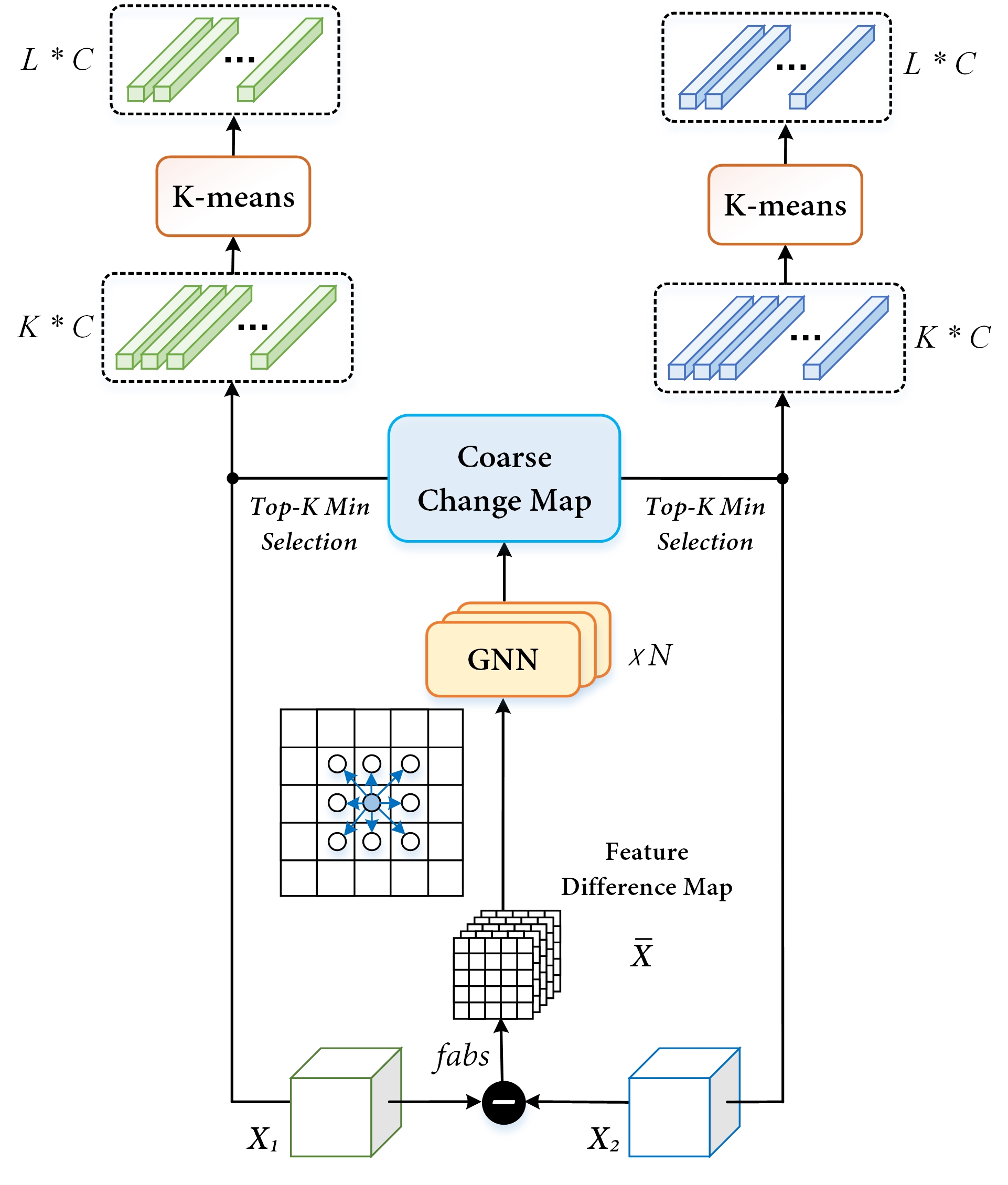} 
\caption{Illustration of our proposed Reliable Token Mining (RTM) module.}
\label{RTM_module}
\end{figure}

To be specific, as illustrated in Fig.~\ref{RTM_module}, the two feature maps are firstly subtracted and we take the absolute values to get the feature difference map $\bar{X}  = \left | X_{1}-X_{2} \right |\in \mathbb{R}^{H\times W\times C}$. First, we build an undirected weight graph $G = \left \{ V, E \right \}$ by treating each feature point (token) as the graph node $v_{i}$ and treating the spatial relationship between node $i$ and $j$ as the edge $e_{i,j}=\left ( v_{i}, v_{j}\right )\in E$. 
Then, based on this graph building, \textbf{our token selection task can be regarded as node selection in graph $G$}.  We employ the GCN to learn the structured information based on such graph $G$ and obtain reliable confidence of graph nodes for node selection. 
Specifically, we first compute the adjacency matrix $A \in \mathbb{R}^{HW\times HW}$ which measures the interactions between node pairs in the graph as 
\begin{equation}
A_{i,j}=
\begin{cases}
\bar{x}_{i}\cdot \bar{x}_{j} &  \quad \text{if }~~~v_{i}, v_{j} ~~~ \text{are adjacent}\\
	0 &       \quad  \text{otherwise}\\
\end{cases}
\end{equation}
where $\bar{x}_i, \bar{x}_j\in \bar{X}$ are feature descriptors for node $v_i$ and $v_j$ respectively. 
The structured information can be modeled and propagated through the graph via GCN module. For the computation defined in each layer $l$ in the GCN, we can formulate it as follows, 
\begin{equation}
H^{\left ( l+1 \right )}=\sigma \left (\widetilde{D}^{-\frac{1}{2}} \big (I+{A}\big )\widetilde{D}^{-\frac{1}{2}} \right )H^{\left(l\right)}W^{\left(l\right)}
\end{equation}
where $I$ is the identity matrix  and $\widetilde{D}$ is the diagonal matrix with $\widetilde{D}_{ii}=\Sigma_j \widetilde{A}_{ij}$ and 
$\widetilde{A}=A+I$. 
$W^{\left(l\right)}$ denotes the learnable parameters. Note that, the initial $H$ in the first layer is set as $H^{(1)}=\bar{X}$ (reshape to $\bar{X}\in \mathbb{R}^{HW\times C} $). After the GCN layers are processed, the final output $P =H^{(L)}\in \mathbb{R}^{H\times W\times 1}$ shares the same spatial resolution as $\bar{X}$. Each element in the final output $P$ corresponds to coarse change confidence. $P$ can be viewed as a one-dimensional coarse probability map or change map, and we show the visualization of $P$ in Section~\ref{SECvisualization}. Obviously, the larger the value of the feature point, the greater the probability that this is a region of change. 
To select the feature tokens with high confidence, we record their position coordinates in the $P$ and get the top-$K$ minima. 
The corresponding selected $K$ features $F_{1} \in \mathbb{R}^{K\times C}$ and $F_{2} \in \mathbb{R}^{K\times C}$ are determined for the dual branches by using the confidence map $P$. 
Reliable tokens are derived from the feature representing the unchanged region on the original feature map, two feature maps construct two sets of tokens from the same region.
To further reduce the token number, we finally utilize the K-means algorithm~\cite{likas2003global} on $F_1$ and $F_2$ to obtain class center $L  (L\ll K)$ centered anchor tokens as $T_{1}, T_{2} \in \mathbb{R}^{L \times C}$ for two branches respectively. $L$ is the length of each set of tokens and $C$ is the channel dimension.

\emph{Discussion:} For the token selection from top-$K$ to $L$, a more simple and intuitive way is to directly choose the $L$ when getting the top-K minima. However, the unchanged areas tend to be far more than the changed areas. Such a naive token selection strategy may be sub-optimal because these tokens may belong to a local region instead of diverse and global regions. In contrast, our proposed two-stage selection way enables \emph{diverse} and \emph{accurate}  selection through large-scale selection and feature clustering in the first and second phase, respectively. 



\noindent  
\textbf{Self-Attention Module. } 
After we get the anchor tokens $T_{1}, T_{2}$ from the above RTM module, we concatenate them together as $T \in \mathbb{R}^{2N \times C} = T_{1} || T_{2}$ and feed them into the self-attention module. Here, the $||$ denotes the concatenate operation. This will enhance the global feature representation and model the relations between different tokens from $T$ mainly due to the computation of the affinity matrix in self-attention. 
To be specific, the standard Transformer block from pre-trained ViT~\cite{dosovitskiy2020image} is adopted to achieve this purpose. It mainly contains positional embedding (PE), prenorm residual unit (PreNorm)~\cite{dosovitskiy2020image}, multi-head self-attention (MSA), and multi-layer perceptron (MLP) block. Before feeding the tokens $T$, we first transform them into query $Q$, key $K$, and value $V$ by using learnable matrices $W^{q} \in \mathbb{R}^{C\times d}, W^{k} \in \mathbb{R}^{C\times d}, W^{v} \in \mathbb{R}^{C\times d}$, 
where $d$ is the channel dimension of $K$, $Q$ and $V$. 
Then, we compute the self-attention in each head  as 
\begin{equation}
\begin{aligned}
& SA(Q, K, V) = Softmax \left ( \frac{QK^{T}}{\sqrt{d_{k}}} \right )V 
\end{aligned}
\end{equation}
The multi-head self-attention (MSA) is utilized by concatenating the learning results of multiple different self-attention modules. The output is fed into the MLP after residual connection and normalization operations. In our implementation, the MLP block consists of two linear layers, and the activation function is the widely used Gaussian Error Linear Unit (GELU)~\cite{hendrycks2016gaussian}. The output of MSA can be split into two parts, i.e., the $T^{*}_{1}$ and $T^{*}_{2}$, for the following cross-attention module which will be introduced below.




\noindent 
\textbf{Cross-Attention Module.} 
The aforementioned self-attention module captures the intra-relationship of given features. In this section, we will introduce the Cross-Attention (CA) module for inter-relation learning between dual anchor token inputs to achieve the information communication between two images. 
Different from the SA module, we first obtain the query $Q$, key $K$, and value $V$ as  
\begin{equation}
\begin{aligned}
Q = T^{*}_{1}, ~~~K = T^{*}_{2}, ~~~V = T^{*}_{2}\\
\end{aligned}
\end{equation} 
Therefore, the cross-attention procedure can be written as: 
\begin{equation}
\begin{aligned}
& CA(Q, K, V) = Softmax \left ( \frac{QK^{T}}{\sqrt{d_{k}}} \right )V 
\end{aligned}
\end{equation}
Two sets of tokens $\widetilde{T}_{1} \in \mathbb{R}^{k \times C}, \widetilde{T}_{2} \in \mathbb{R}^{k \times C}$ are obtained from cross-attention for dual images and are fed into the followed Anchor-Primary Attention for the final change detection.





\noindent 
\textbf{Anchor-Primary Attention.}
In this section, we employ Siamese anchor-primary attention to obtain the final feature maps by refining features in pixel-level space. The architecture of the anchor-primary attention block is similar to the aforementioned Transformer block, but without the PE block, as similarly suggested in works~\cite{lee2019set,jiang2023amatformer}. It mainly consists of PreNorm, Multi-Head Anchor-Primary Attention (MAPA), and MLP. To be specific, in our anchor-primary attention, key $K$ and value $V$ are obtained from the tokens $\widetilde{T}_{1}$ or $\widetilde{T}_{2}$, while the query $Q$ is obtained from the original feature maps. 
Formally, the Anchor-Primary Attention (APA) can be formulated as 
\begin{equation}
\begin{aligned}
&\widetilde{X}_i =APA(Q,K,V) \\
& = APA(X_i  W^{q}, \widetilde{T}_{i} W^{k}, \widetilde {T}_{i}  W^{v}) 
\end{aligned}
\end{equation}
\begin{equation}
\begin{aligned}
& APA(Q, K, V) = Softmax \left ( \frac{QK^{T}}{\sqrt{d_{k}}} \right )V 
\end{aligned}
\end{equation}
where $i=\{1, 2\}$ and $W^{q}, W^{k}, W^{v} \in \mathbb{R}^{C \times d}$ are learnable parameters. 

\noindent 
\textbf{Prediction Head.} 
Once we obtain the enhanced features from the above anchor-primary attention module, we first reshape the feature vectors into 2D maps $X'_{1}, X'_{2} \in \mathbb{R}^{H \times W \times C}$. Then, a prediction head is proposed to transform the features into the final change map results. Specifically, the  2D feature maps are subtracted  to produce the feature-level difference maps $D = \big | {X'_{1}-X'_{2} \big |}$. It is then upsampled to the scale of the original image and fed into the convolutional neural networks to obtain the predicted map  $\mathcal{P} \in \mathbb{R}^{H_{0} \times W_{0} \times 2}$. 




\subsection{Loss Function} 
The change detection is formulated as a binary classification problem, and the cross-entropy loss function is used for the training of our proposed VcT method. Note that, our model outputs a change map with two dimensions. The first dimension denotes the probability/confidence of unchanged regions, while the second dimension represents the changed regions. The ground truth is expanded from one channel to two channels, with one-hot encoding for each pixel $\mathcal{G} \in \mathbb{R}^{H_{0}\times W_{0}\times 2}$. The loss is calculated with outputs and the one-hot encoding of ground truth, i.e., 
$$L_{bce}(\mathcal{G},\mathcal{P})=-\frac{1}{H_{0}\times W_{0}}\sum_{i=1}^{H_{0}\times W_{0}}\mathcal{G}(i)\log \mathcal{P}\left ( i \right )$$ 
where $\mathcal{G}$ represents the true ground truth value and $\mathcal{P}$ denotes the predicted value. The $H_0$ and $W_0$ denote the height and width of input images respectively.


\section{Experiments} \label{experiments}

\begin{table*}[t]
\caption{Comparisons with other SOTA models on three remote sensing change detection datasets. The best and second results are marked in \textcolor{black}{RED} and \textcolor{blue}{BLUE}, respectively. All these scores are written in percentage (\%).} 
\label{benchmarkResults}
\centering
\resizebox{\textwidth}{22mm}{
\begin{tabular}{lccccccccccccccccc}
\toprule
\textbf{Method} & \multicolumn{5}{c}{\textbf{LEVIR-CD}~\cite{chen2020spatial}} && \multicolumn{5}{c}{\textbf{WHU-CD}~\cite{ji2018fully}} && \multicolumn{5}{c}{\textbf{DSIFN-CD}~\cite{zhang2020deeply}}\\
\cmidrule{2-6} \cmidrule{8-12} \cmidrule{14-18}
&Pre. & Rec. & F1 & IoU & OA && Pre. & Rec. & F1 & IoU & OA  && Pre. & Rec. & F1 & IoU & OA \\
\toprule
\textbf{FC-EF}~\cite{daudt2018fully}        &86.91 & 80.17 & 83.40 & 71.53 & 98.39 && 71.63 & 67.25 & 69.37 & 53.11 & 97.61 && 72.61 & 52.73 & 61.09 & 43.98 & 88.59\\
\textbf{FC-Siam-Di}~\cite{daudt2018fully}   &89.53 & 83.31 & 86.31 & 75.92 & 98.67 && 47.33 & 77.66 & 58.81 & 41.66 & 95.63 && 59.67 & 65.71 & 62.54 & 45.50 & 86.63\\
\textbf{FC-Siam-Conc}~\cite{daudt2018fully} &91.99 & 76.77 & 83.69 & 71.96 & 98.49 && 60.88 & 73.58 & 66.63 & 49.95 & 97.04 && 66.45 & 54.21 & 59.71 & 42.56 & 87.57\\
\textbf{DTCDSCN}~\cite{liu2020building}      &88.53 & 86.83 & 87.67 & 78.05 & 98.77 && 63.92 & 82.30 & 71.95 & 56.19 & 97.42 && 53.87 & \textcolor{black}{\bf77.99} & 63.72 & 46.76 & 84.91\\
\textbf{STANet}~\cite{chen2020spatial}       &83.81 &\textcolor{black}{\bf 91.00} & 87.26 & 77.40 & 98.66 && 79.37 & \textcolor{blue}{\bf85.50} & 82.32 & 69.95 & 98.52 && 67.71 & 61.68 & 64.56 & 47.66 & 88.49\\
\textbf{IFNet}~\cite{zhang2020deeply}        &\textcolor{black}{\bf 94.02} & 82.93 & 88.13 & 78.77 & 98.87 && 78.00 & 70.81 & 74.23 & 59.03 & 92.53 && 67.86 & 53.94 & 60.10 & 42.96 & 87.83 \\
\textbf{SNUNet}~\cite{fang2021snunet}       & 89.18 & 87.17 & 88.16 & 78.83 & 98.82 && 85.60 & 81.49 & 83.50 & 71.67 & 98.71 && 60.60 & \textcolor{blue}{\bf72.89} & 66.18 & 49.45 & 87.34\\
\textbf{CropLand}~\cite{liu2022cnn}       & 89.79 & 87.57 & 88.67 & 79.64 & 98.86 && 83.87 & 75.81 & 79.64 & 66.17 & 94.11 && 61.72 & 65.08 & 60.53 & 43.40 & 87.03\\
\textbf{DMATNet}~\cite{song2022remote}       & 91.56 & \textcolor{blue}{\bf89.98} & \textcolor{black}{\bf90.75} & \textcolor{black}{\bf84.13} & 98.25 && \textcolor{black}{\bf89.46} & 82.24 & \textcolor{blue}{\bf85.70} & \textcolor{blue}{\bf74.98} & 95.83 && 66.65 & 76.50 & 71.23 & 55.32 & 87.12\\
\hline 
\textbf{BIT}~\cite{chen2021remote} & 89.24 & 89.37 & 89.31 & 80.68 & \textcolor{blue}{\bf98.92} && 86.64 & 81.48 & 83.98 & 72.39 & \textcolor{blue}{\bf98.75} && \textcolor{black}{\bf86.28} & 61.56 & \textcolor{blue}{\bf71.85} & \textcolor{blue}{\bf56.07} & \textcolor{blue}{\bf91.81}\\
\textbf{VcT (Ours)}     & \textcolor{blue}{\bf92.57} & 87.65 & \textcolor{blue}{\bf90.04} & \textcolor{blue}{\bf81.89} & \textcolor{black}{\bf99.01} && \textcolor{blue}{\bf89.39} & \textcolor{black}{\bf89.77} & \textcolor{black}{\bf89.58} & \textcolor{black}{\bf81.12} & \textcolor{black}{\bf99.18} && \textcolor{blue}{\bf83.91} & 66.47 & \textcolor{black}{\bf74.18} & \textcolor{black}{\bf58.95} & \textcolor{black}{\bf92.14} \\
\bottomrule
\end{tabular}}
\end{table*}

\begin{table}[t]
\caption{Ablation study of core components of our proposed VcT on the LEVIR-CD dataset. All these scores are written in percentage (\%).} 
\label{DifferentComponentsAnalysis}  
\centering 
\resizebox{0.48\textwidth}{11mm}{ 
\begin{tabular}{ccccc|ccc}
\toprule
Index &Backbone &RTM  &TE         &TD           & F1\; $|$ IoU $|$ OA  \\
\hline 
1 &\textcolor{SeaGreen4}{\cmark}   &\textcolor{DarkRed}{\xmark}   &\textcolor{SeaGreen4}{\cmark}     &\textcolor{SeaGreen4}{\cmark}     &    89.09$|$80.33$|$98.93  \\
2 &\textcolor{SeaGreen4}{\cmark}   &\textcolor{SeaGreen4}{\cmark}   &\textcolor{DarkRed}{\xmark}     &\textcolor{SeaGreen4}{\cmark}       & 88.39$|$79.20$|$98.88  \\
3 &\textcolor{SeaGreen4}{\cmark}   &\textcolor{SeaGreen4}{\cmark}   &\textcolor{SeaGreen4}{\cmark}     &\textcolor{DarkRed}{\xmark}       & 89.37$|$80.78$|$98.94 \\
4 &\textcolor{SeaGreen4}{\cmark}   &\textcolor{SeaGreen4}{\cmark}   &\textcolor{SeaGreen4}{\cmark}     &\textcolor{SeaGreen4}{\cmark}     & 90.04$|$81.89$|$99.01  \\
\bottomrule
\end{tabular} } 
\end{table}

\begin{table}[t]
\caption{Ablation study of GNN and K-means in our proposed RTM module on the LEVIR-CD dataset. All these scores are written in percentage (\%).}
\label{AblationStudyonGNNandK-means}  
\centering 
\begin{tabular}{ccc|ccc}
\toprule
Index  &GNN      &K-means           & F1\, $|$ IoU $|$ OA  \\
\hline 
1  &\xmark       &\cmark     & 89.81$|$81.51$|$98.99  \\
2  &\cmark       &\xmark     & 88.47$|$79.32$|$98.89 \\
3  &\cmark       &\cmark     & \textbf{90.04}$|$\textbf{81.89}$|$\textbf{99.01} \\ 
\bottomrule
\end{tabular} 
\end{table}

\subsection{Dataset and Evaluation Metric}   
In our experiments, three widely used HSR remote sensing image datasets are used, including \textbf{LEVIR-CD}~\cite{chen2020spatial}, \textbf{WHU-CD}~\cite{ji2018fully}, and \textbf{DSIFN-CD}~\cite{zhang2020deeply}. A brief introduction to these datasets is given below.

$\bullet$ \textbf{LEVIR-CD}~\cite{chen2020spatial} is a remote sensing dataset specifically designed for building change detection. It consists of 637 image patch pairs of very high resolution (VHR) with a resolution of 0.5m/pixel and size of $1024 \times 1024$ pixels, obtained from Google Earth. The dataset is annotated by experts and contains a total of 31,333 individual examples of changing buildings.  Each image pair is divided into non-overlapping patches of size  $256 \times 256$ pixels. The dataset is further split into training, validation, and testing subsets, with 7120, 1024, and 2048 image pairs, respectively.

$\bullet$ \textbf{WHU-CD}~\cite{ji2018fully} The dataset documents the changes in the affected area after the 6.3 magnitude earthquake and the reconstruction a few years later, taken in 2012 and 2016, respectively, and contains more than 10,000 buildings within 20.5 square kilometers. The dataset was geo-corrected to 1.6 pixel accuracy for the aerial dataset.
Each image has a spatial size of 15354×32507 pixels with a spatial resolution of 0.2m. We divide each image into nonoverlapping patches of size 256 × 256. Therefore, we obtain training, validation, and testing subset containing 6096, 760, and 760 image pairs, respectively.

$\bullet$ \textbf{DSIFN-CD}~\cite{zhang2020deeply} is a dataset for building change detection consisting of six large diachronic high-resolution images covering six cities in China, including Beijing, Chengdu, Shenzhen, Chongqing, Wuhan, and Xi’an. The images were manually collected from Google Earth and were cropped into 394 sub-image pairs of size $512 \times 512$. After data augmentation, a total of 3940 dual-temporal image pairs were obtained. The remaining image pairs were cropped into 48 pairs for model testing. Non-overlapping patches of size $256 \times 256$ were created by slicing the $512 \times 512$ image, in line with some of the latest change detection methods, while utilizing the authors’ default training/validation/testing sets. The dataset contains 14400, 1360, and 192 image pairs in the training, validation, and testing subsets, respectively.

In our experiments, we use five evaluation metrics to assess the performance of change detection algorithms. These metrics include \textbf{Precision}, \textbf{Recall}, \textbf{IoU} (Intersection over Union), and \textbf{OA} (Overall Accuracy)~\cite{chen2021remote}, which are defined as follows:
\begin{flalign}  
& Precision = TP/(TP + FP)  \\
& Recall = TP/(TP+FN)   \\
& IoU = TP/(TP+FN+FP)   \\
& OA = (TP+TN)/(TP+TN+FN+FP) 
\end{flalign}
where TP, TN, FP, and FN represent the number of true positive, true negative, false positive, and false negative, respectively.
In particular, the F1-score takes into account both the Precision and Recall of the classification model~\cite{chen2021remote}. We use it with regard to the change category as the main evaluation. 

\subsection{Implementation Details} 
Our proposed VcT framework is trained end-to-end using SGD~\cite{sutskever2013SGD} optimizer with a linear learning rate policy. The model is trained for 200 epochs with an initial learning rate of 0.01, batch size of 8, weight decay of 0.0005, and momentum of 0.99. The reliable token mining (RTM) module is fine-tuned by testing different parameters for $K$, $L$, and the number of GNN layers $N$. The final values are set to $K=1000$, $L=10$, $N=1$, while the 8-nearest neighbor graph is used. In Transformer layers, the number of heads in MSA and MAPA is set to 8. Our model is implemented in Python using the PyTorch~\cite{paszke2019pytorch} toolkit and trained on a server equipped with a NVIDIA GeForce RTX 3090 GPU.


\begin{figure} [t]
\centering
\includegraphics[width=0.5\textwidth]{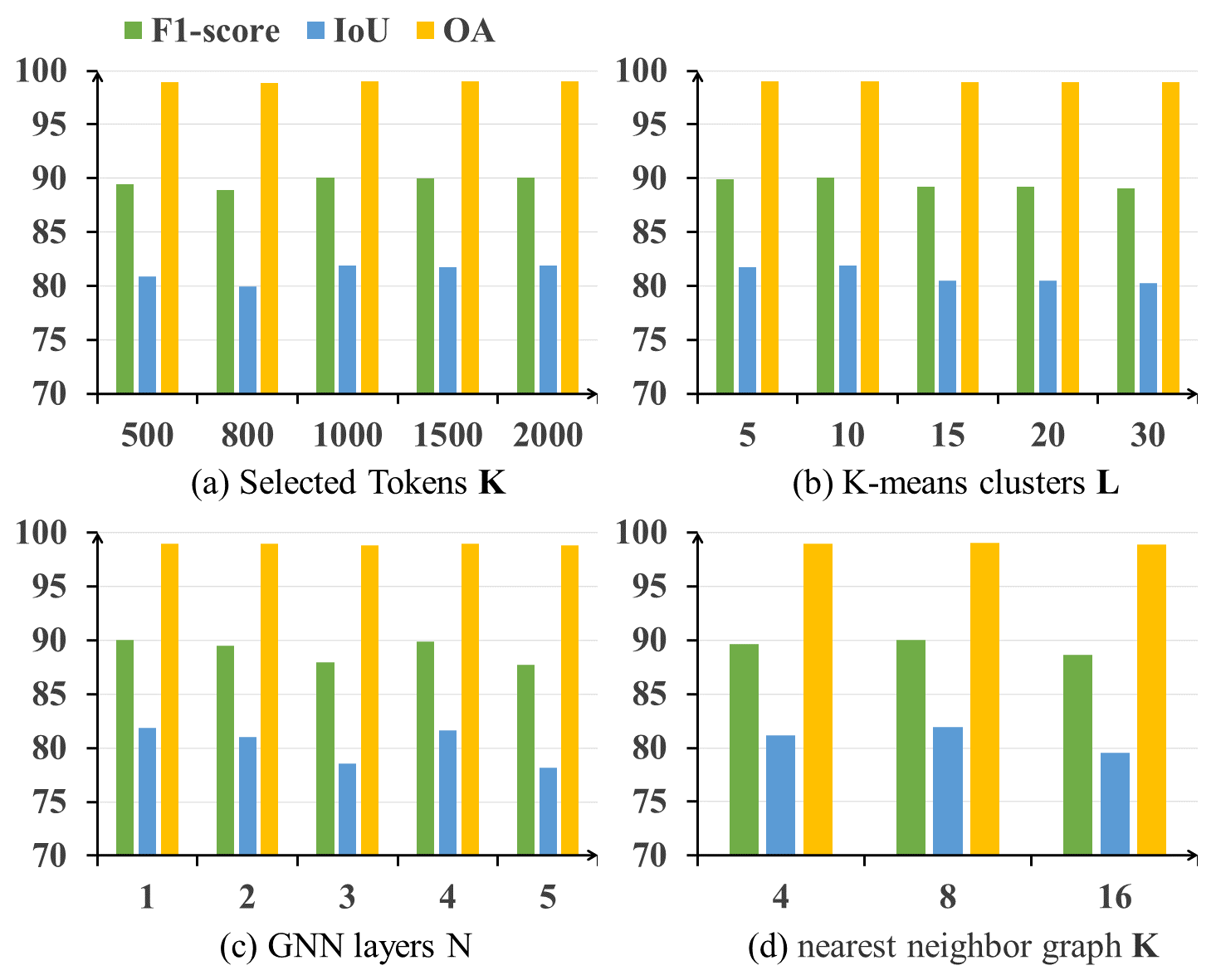} 
\caption{Analysis of selected tokens, K-means clusters, GNN layers and different nearest neighbors on the LEVIR dataset. } 
\label{Histogram}
\end{figure}

\begin{figure*}[t]
\centering
\includegraphics[width=1\textwidth]{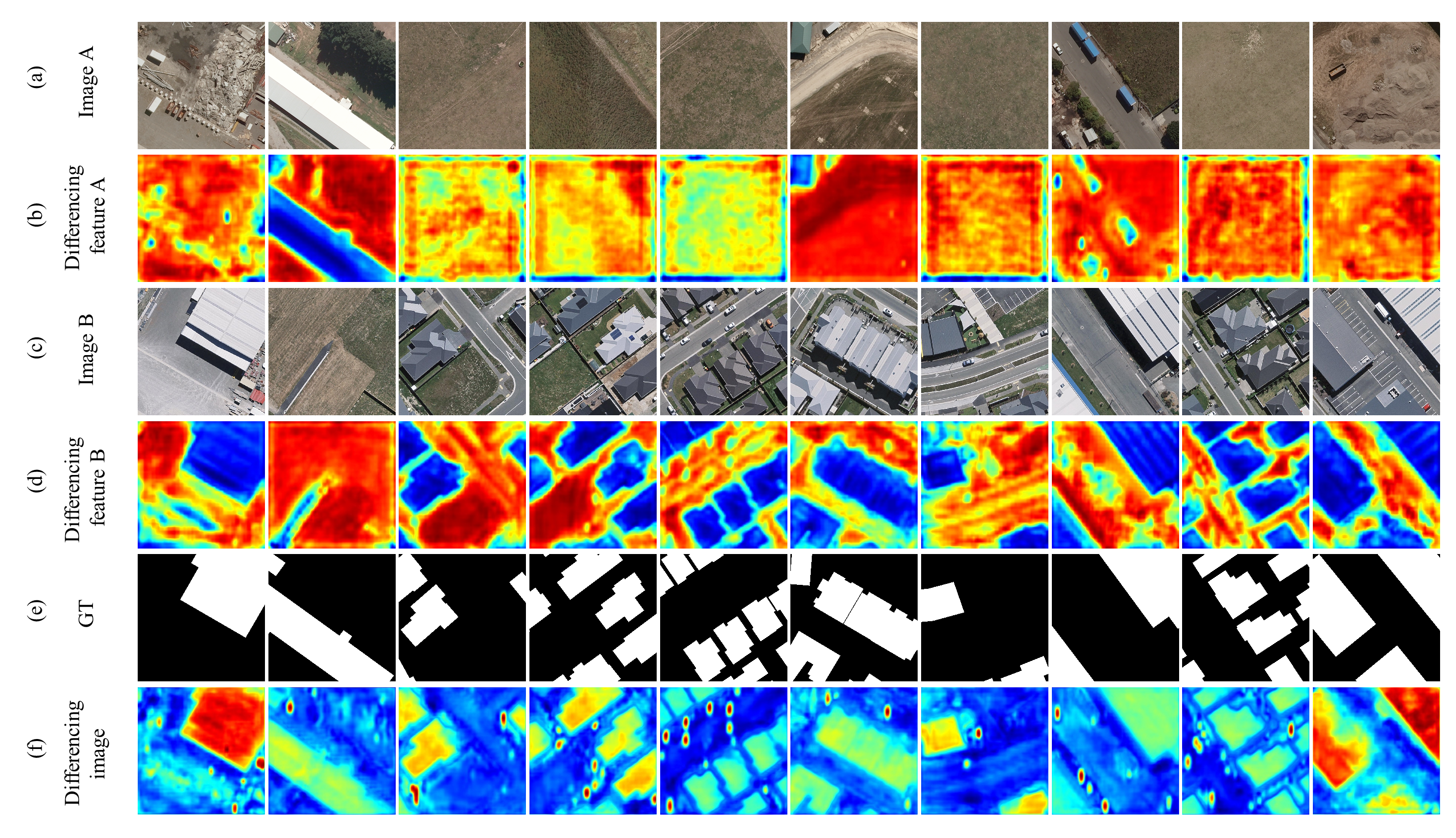} 
\caption{Example of feature maps visualization on WHU-CD test set. Red and blue denotes higher and lower attention values respectively. (a) Image A, (b) Differencing feature map of image A, (c) Image B, (d) Differencing  feature map of image B, (e) Ground Truth, (f) Differencing image.}
\label{visFeatureMaps}
\end{figure*}

\begin{figure*}[htp]
\centering
\includegraphics[width=1\textwidth]{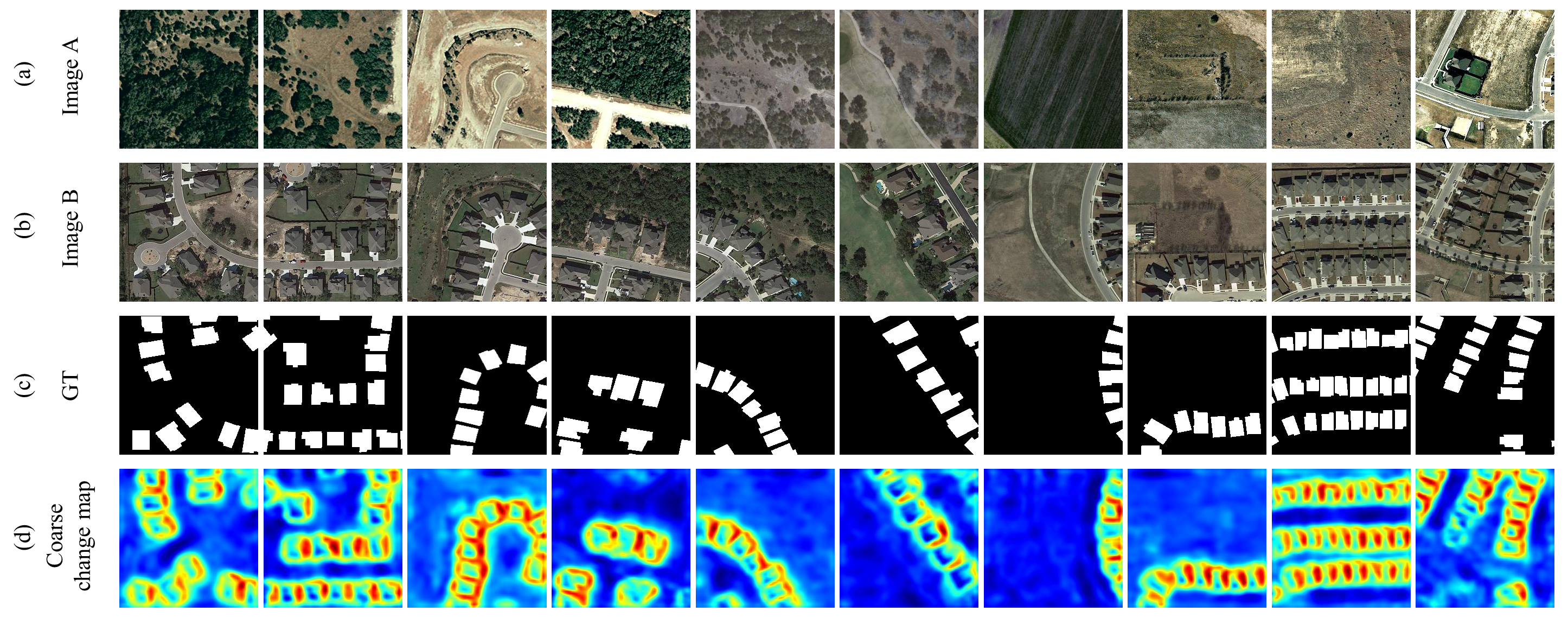} 
\caption{Visualization of representative coarse change map on the LEVIR-CD test set.} 
\label{visCoarseChangeMap}
\end{figure*}

\begin{figure*}[htp]
\centering
\includegraphics[width=1\textwidth]{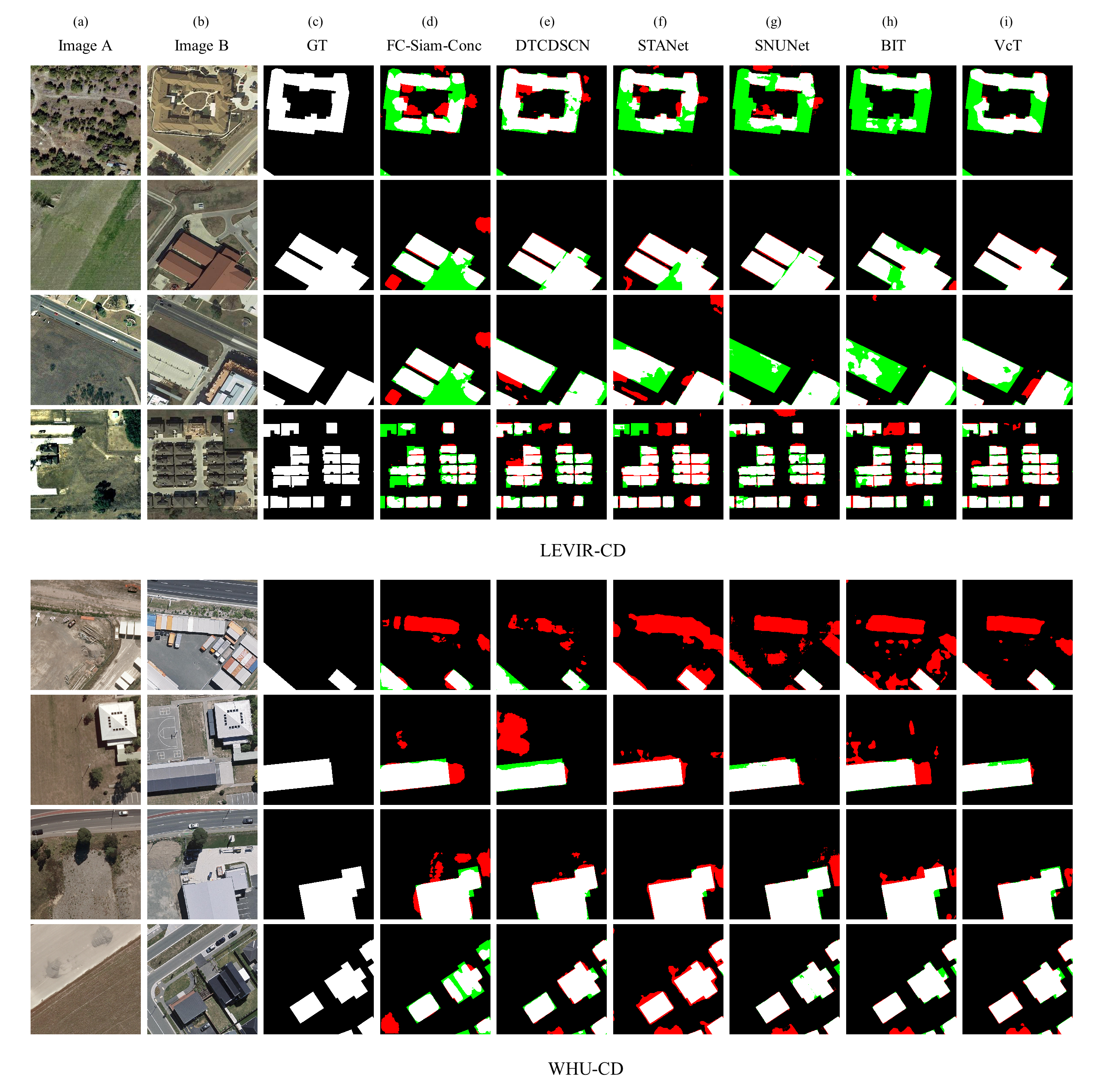} 
\caption{Visualization of change detection results of our proposed VcT and other SOTA models.} 
\label{visDetectionResults}
\end{figure*}

\subsection{Comparison with State-Of-The-Art Models}  

As mentioned in previous sections, we validate our proposed method on three benchmark datasets and compare our method with 10 state-of-the-art change detection models, including FC-EF~\cite{daudt2018fully}, FC-Siam-Di~\cite{daudt2018fully}, FC-Siam-Conc~\cite{daudt2018fully}, DTCDSCN~\cite{liu2020building}, STANet~\cite{chen2020spatial}, IFNet~\cite{zhang2020deeply}, SNUNet~\cite{fang2021snunet}, BIT~\cite{chen2021remote}, CropLand~\cite{liu2022cnn}, DMATNet~\cite{song2022remote}. 
Note that the first four methods~\cite{daudt2018fully} are based on purely convolutional neural network architectures, and the remaining six models are based on Transformer methods. 
The experimental results reported in Table~\ref{benchmarkResults} are implemented based on their source codes and default parameters. More detailed results and analyses of these datasets are given below. 

\begin{enumerate}
\item[1)] FC-EF~\cite{daudt2018fully}: The method is a single-stream network, where two images are concatenated as a single input and fed into a full convolutional network (FCN).
\textcolor{black}{The model uses the SGD optimizer with a learning rate of 0.001, momentum of 0.9, and weight decay of 0.0005. The batchsize is set to 10.}

\item[2)] FC-Siam-Di~\cite{daudt2018fully}: The method is a dual-stream network, where two images are extracted features by using two FCN encoders, and the difference operation is first performed on the two image features, and the extracted difference features at different levels are input to the FCN decoder.
\textcolor{black}{The parameters setting of this method are the same as FC-EF.}

\item[3)]FC-Siam-Conc~\cite{daudt2018fully}: The method is a dual-stream network, where two images are extracted by two FCN encoders respectively, and the features are concatenated together and input to a FCN decoder.
\textcolor{black}{The parameters setting of this method are the same as FC-EF.}

\item[4)]DTCDSCN~\cite{liu2020building}: The method is a dual-stream network that introduces spatial attention and channel attention in the FCN, thus improving the feature representation. The method contains three sub-networks, i.e., one change detection network and two semantic segmentation networks. Similar to BIT~\cite{chen2021remote}, we omit the semantic segmentation decoders for the fair comparison.
\textcolor{black}{The experiment uses the small batch ADAM algorithm to train the network. The batch size is set to 16 and the initial learning rate is set to 0.001.}

\item[5)]STANet~\cite{chen2020spatial}: The method is a dual-stream network, where the features of two images are extracted by using two encoders, together with the spatial-temporal attention mechanism.
\textcolor{black}{The model use Adam solver with a batch size of 4 and an initial learning rate of 0.001. It keep the same learning rate for the first 100 epochs and linearly decay the learning rate to 0 for the remaining 100 epochs.}

\item[6)] IFNet~\cite{zhang2020deeply}: The method is a dual-stream network that extracts features from the image via FCN dual-stream structure, and then the extracted deep features are fed into a deeply supervised difference discrimination network (DDN) for change detection.
\textcolor{black}{The learning rate is set to 0.0001 and decreased by 10\% when the loss stops decreasing for 5 epochs. Model training ends when the score of f1 on the validation dataset does not improve for 20 epochs.}

\item[7)] SNUNet~\cite{fang2021snunet}: The method is a single-stream network, which uses a combination of the Siamese network and NestedUNet~\cite{zhou2018unet++}, detected by an encoder and decoder containing an Ensemble Channel Attention Module (ECAM).
\textcolor{black}{The experiment batch size is set to 16, and Adam is used as an optimizer. The learning rate is set to 0.001 and decays by 0.5 every 8 epochs until 200 epochs.}

\item[8)] BIT~\cite{chen2021remote}: The method is a dual-stream network, which extracts high-level features via convolutional networks and constructs semantic tokens by using a Transformer.
\textcolor{black}{The learning rate, weight decay and momentum are set to 0.01, 0.0005 and 0.99 respectively.}

\item[9)] CropLand~\cite{liu2022cnn}: The method is a single-stream network, which first extracts multi-scale features by using CNNs and designs a transformer-based MSCA to encode and aggregate contextual information.
\textcolor{black}{The experiment optimized the model using 8 batches size and an Adam optimizer with 0.0001 learning rate, training process lasts for 100 epochs.}

\item[10)] DMATNet~\cite{song2022remote}: The method is a dual-stream network, which uses a dual feature extraction method with a dual feature mixture attention (DFMA) module to fuse fine and coarse features.
\textcolor{black}{The model is optimized by using SGD algorithm. The momentum is set to 0.99 and the weight attenuation is set to 0.0005. The learning rates are set to 0.01, 0.0006, and 0.01, respectively for LEVIR-CD, DSIFN-CD and WHU-CD datasets.}
\end{enumerate}

\noindent 
\textbf{Results on LEVIR-CD Dataset}~\cite{chen2020spatial}. 
As shown in Table~\ref{benchmarkResults}, our baseline method BIT~\cite{chen2021remote} achieves $89.24\%$, $89.37\%$, $89.31\%$, $80.68\%$, $98.92\%$ on the Precision, Recall, F1-score, IoU, and OA metric, respectively. In contrast, our proposed VcT obtains $92.57\%$, $87.65\%$, $90.04\%$, $81.89\%$, $99.01\%$, which outperforms the BIT model on most of these metrics. Specifically, we beat the BIT on Precision, F1-score, IoU, and OA by $+2.76\%$, $+1.02\%$, $+1.7\%$, $+0.11\%$ respectively. These experimental results show the effectiveness of our proposed VcT framework for remote sensing image change detection task. It is easy to find that our proposed framework obtains improved results than other Transformer based change detection algorithms, such as DTCDSCN~\cite{liu2020building}, STANet~\cite{chen2020spatial}, IFNet~\cite{zhang2020deeply}, SNUNet~\cite{fang2021snunet}, etc. These results fully demonstrate the advantages and superior performance of the proposed VcT model.

\noindent 
\textbf{Results on WHU-CD Dataset}~\cite{ji2018fully}.  
According to the results of WHU-CD dataset reported in Table~\ref{benchmarkResults}, we can find that the proposed VcT achieves $89.39\%$/$89.77\%$/$89.58\%$, $81.12\%$, $99.18\%$ on the 
P/R/F1, IoU, and OA metric, respectively. Compared to baseline method BIT~\cite{chen2021remote} which obtains $86.64\%$, $81.48\%$, $83.98\%$, $72.39\%$, $98.75\%$, VcT has improved all the five evaluation indicators by $+2.75\%$, $+8.29\%$, $+5.6\%$, $+8.73\%$, $+0.43\%$ respectively. We can also find that our model obtains better results than other change detection algorithms. 

\noindent 
\textbf{Results on DSIFN-CD Dataset}~\cite{zhang2020deeply}. 
From the Table~\ref{benchmarkResults}, it can be concluded that the proposed VcT performs better than the baseline BIT~\cite{chen2021remote} in multiple metrics on this dataset. Specifically, we beat the BIT on Recall, F1-score, IoU, and OA by $+4.91\%$, $+2.33\%$, $+2.88\%$, $+0.33\%$ respectively. \textcolor{black}{Since the DSIFN-CD dataset is challenging and it is usually difficult to detect the changed regions accurately, the compared methods generally obtain low Recall. However} we can also find that the proposed VcT model obtains better results than other Transformer-based change detection algorithms, such as DTCDSCN~\cite{liu2020building}, STANet~\cite{chen2020spatial}, IFNet~\cite{zhang2020deeply}, SNUNet~\cite{fang2021snunet}, CropLand~\cite{liu2022cnn}, DMATNet~\cite{song2022remote}. 

Overall, these experiments fully demonstrate the effectiveness and superiority of our newly proposed VcT for the remote sensing change detection task. 

\begin{table}[t]
\caption{Results of Different Selected Tokens on LEVIR-CD dataset. All these scores are written in percentage (\%).}  
\label{SelectedTokens}  
\small 
\centering 
\begin{tabular}{c|c|c|c|c|c}
\toprule
K          &500 & 800 & 1000 & 1500& 2000 \\
\hline
\rule{0pt}{8pt}
F1   & 89.41&88.88&\textbf{90.04}&89.96&90.03 \\ 
IoU  & 80.85&79.99&\textbf{81.89}&81.76&81.87 \\ 
OA   & 98.95&98.88&\textbf{99.01}&98.99&\textbf{99.01}\\
\bottomrule
\end{tabular} 
\end{table}

\begin{table}[t]
\caption{Results of Various Clusters on LEVIR-CD dataset. All these scores are written in percentage (\%).}  
\label{Clusters}  
\centering 
\begin{tabular}{c|c|c|c|c|c}
\toprule
L          &5 & 10 & 15 & 20 & 30 \\
\hline 
\rule{0pt}{8pt}
F1   & 89.93&\textbf{90.04}&89.21&89.21&89.06 \\ 
IoU  & 81.70&\textbf{81.89}&80.52&80.51&80.28 \\ 
OA   & 98.99&\textbf{99.01}&98.94&98.94&98.92 \\
\bottomrule
\end{tabular} 
\end{table}

\begin{table}[t]
\caption{Effects of Different GNN Layers on the LEVIR-CD dataset. All these scores are written in percentage (\%).}  
\label{GNNlayers}  
\centering 
\begin{tabular}{c|c|c|c|c|c}
\toprule
n          &1 &2 &3 &4 &5  \\
\hline 
\rule{0pt}{8pt}
F1   & \textbf{90.04}&89.52&87.99&89.87&87.73 \\ 
IoU  & \textbf{81.89}&81.02&78.56&81.61&78.16 \\ 
OA   & \textbf{99.01}&98.96&98.84&98.99&98.83 \\
\bottomrule
\end{tabular} 
\end{table}

\begin{table}[t]
\caption{Analysis on Different Nearest Neighbors on the LEVIR-CD dataset. All these scores are written in percentage (\%).}  
\label{NearestNeighbors}  
\centering 
\begin{tabular}{c|c|c|c}
\toprule
k-nn          &4-nn &8-nn &16-nn  \\
\hline
\rule{0pt}{8pt}
F1   & 89.61&\textbf{90.04}&88.62 \\ 
IoU  & 81.17&\textbf{81.89}&79.56 \\ 
OA   & 98.98&\textbf{99.01}&98.87 \\
\bottomrule
\end{tabular} 
\end{table}

\subsection{Ablation Study} 
In this subsection, we conduct the following ablation studies to  better understand our key contributions, including different components analysis,  number of selected tokens,  number of clusters, GNN layers, different nearest neighbors, etc.

\noindent 
\textbf{Different Components Analysis.} In the proposed VcT, there are four main modules including the shared backbone network, RTM module, Self-/Cross-Attention module, and Anchor-Primary Attention module. We use Self-/Cross-Attention module as Transformer encode (TE) and Anchor-Primary Attention module as Transformer Decoder (TD). As shown in Table~\ref{DifferentComponentsAnalysis}, we remove each of these components gradually to check their influence on final detection results on the LEVIR-CD dataset, i.e., from algorithm 1 to algorithm 4. 
We can see that the best performance can be achieved when all the components are used. 
To be specific, the performance of our proposed VcT without the proposed RTM module are reduced to $89.09\%$, $80.33\%$, $98.93\%$, which validates the effectiveness and importance of RTM module for the proposed VcT framework. 
In addition, the performance of our model are dropped to $88.39\%$, $79.20\%$, $98.88\%$ when the TE is removed and $89.37\%$, $80.78\%$, $98.94\%$ when the TD is removed. 
These results prove the effectiveness of Transformer network for the proposed VcT framework. 
In conclusion, these experimental results fully demonstrate each key component contributes to our VcT framework.


\noindent 
\textbf{Ablation Study on GNN and K-means.}
In this subsection, we conduct the following analysis to help readers better understand our Reliable Token Mining (RTM) module, to verify the effectiveness of GNN layers and K-means. As shown in Table~\ref{AblationStudyonGNNandK-means}, we remove GNN layers and K-means respectively to check their influence on final detection results on the LEVIR-CD dataset. We can see that the best performance can be achieved when all the components are used. To be specific, the performance of our proposed RTM module replacing GNN with CNN is reduced to 89.81\%, 81.51\%, and 98.99\%, which validates the effectiveness and importance of GNN for the proposed RTM module. In addition, the performance of our model dropped to 88.47\%, 79.32\%, and 98.89\% when the K-means module is removed. These results demonstrate the effectiveness of GNN and K-means in our RTM module. 

\noindent 
\textbf{Effects of the Number of Selected Tokens.} 
The number of selected tokens $K$ plays an important role in our proposed RTM module. 
It makes the changed region be interfering with the context modeling of the common region when the $K$ is too large. 
On the contrary, the utilization of the common region is low and the prior information cannot be fully exploited when the $K$ is too small. 
In this subsection, we test different tokens $K$ to find the tradeoff between these two aspects. As shown in Fig.~\ref{Histogram} (a) and Table~\ref{SelectedTokens}, we set the $K$ ranging from $500$ to $2000$ and conduct the experiments on the LEVIR dataset. We can find that the best results can be obtained when $K=1000$ and thus we set $K$ to 1000.  


\noindent 
\textbf{Effects of the Number of Clusters.}
The K-means clustering algorithm is used in the RTM module.
Here, we set different clustering settings (e.g., $L=\{5, 10, 15, 20, 30\}$) to check their influence on the final results. As shown in Fig.~\ref{Histogram} (b) and Table~\ref{Clusters}, 
we can observe that our results are not sensitive to this parameter. 
Slightly better results can be obtained when the cluster is set as 10.
Thus, we set the number of clusters to 10. 


\noindent 
\textbf{Effects of Different GNN Layers.}
It is known that deeper layers of GNN may lead to the issue of over-smoothing. 
To study the influence of the number of GNN layers in the proposed VcT framework,
we set the GNN layers ranging from 1 to 5 and conduct experiments on LEVIR dataset. 
As shown in Fig.~\ref{Histogram} (c) and Table~\ref{GNNlayers}, we can observe that better performance can be obtained when we just use one GNN layer.


\noindent 
\textbf{Analysis on Different Nearest Neighbors. }
To check the influence of different nearest neighbors for the graph construction, we test 4-NN, 8-NN, and 16-NN. As illustrated in Fig.~\ref{Histogram} (d) and Table~\ref{NearestNeighbors}, we can find that the best performance can be achieved on the LEVIR dataset when the 8-NN graph is used. Thus, we select the 8-NN graph for the graph construction. 

\textcolor{black}{The parameters on other datasets (i.e., WHU-CD, DSIFN-CD) are the same as those of LEVIR dataset. We can find that our results are consistent better than the compared methods and relatively stable, which demonstrate that the optimal parameters on the LEVIR dataset are suitable for other datasets.}

\begin{table}[ht]
\caption{Parameters and running efficiency on the LEVIR-CD dataset. }  
\label{Parameters and FLOPs}  
\centering 
\setlength\tabcolsep{10pt}{
\begin{tabular}{ccc}
\toprule
Model & Params.(M) & FLOPs(G)  \\
\hline 
\rule{0pt}{8pt}
DTCDSCN      & 31.26M & 7.21G  \\
STANet       & 16.89M & 6.58G     \\
IFNet        & 50.71M  & 41.18G     \\
SNUNet       & 12.03M & 27.44G    \\
BIT          & 3.50M  & 10.63G    \\
VcT          & 3.57M  & 10.64G    \\
\bottomrule
\end{tabular} }
\end{table}

\noindent 
\textbf{Parameters and Running Efficiency. } 
To make better understand the efficiency of our model, here, we report the model parameters (Params.) and floating-point operations per second (FLOPs) of our model and five other SOTA methods. All these results are tested on a server with an Intel(R) Xeon(R) Silver 4314 CPU and a GeForce RTX 3090 GPU. 
As shown in Table~\ref{Parameters and FLOPs}, we can see the parameters of our proposed VcT model is 3.57M while DTCDSCN~\cite{liu2020building}, STANet~\cite{chen2020spatial}, IFNet~\cite{zhang2020deeply}, SNUNet~\cite{fang2021snunet} and BIT~\cite{chen2021remote} is 31.26M, 16.89M, 50.71M, 12.03M, 3.50M, respectively. Moreover, the FLOPs of our model is 10.64G, while DTCDSCN, STANet, IFNet, SNUNet, and BIT are 7.21G, 6.58G, 41.18G, 27.44G, and 10.63G, respectively.
It is easy to find that the complexity and efficiency of our model are comparable to the baseline method BIT and obviously better than some other compared   works.

\subsection{Visualization} \label{SECvisualization}
In addition to the aforementioned quantitative analysis, we also give some intuitive examples to better understand our proposed model from the perspective of qualitative analysis. 
To be specific, we conduct the visualization of feature maps, coarse change maps and final detection results.

\noindent 
\textbf{Feature Maps. }   
As shown in Fig.~\ref{visFeatureMaps}, given the input Image A (a) and Image B (c), our proposed RTM module selects the common invariant background regions for fusion. Through Fig.~\ref{visFeatureMaps} (b) and Fig.~\ref{visFeatureMaps} (d) which are the difference feature maps of the enhanced and the original feature map, we can observe that it enhances the background representation and eliminates the irrelevant changes. Therefore, higher-quality changed maps can be detected by using the proposed model, as shown in Fig.~\ref{visFeatureMaps} (f).

\noindent 
\textbf{Coarse Change Map.}
As shown in Fig.~\ref{visCoarseChangeMap}, we give a visualization of the coarse change maps of some representative samples. 
We can find that our proposed RTM module can first roughly capture the common unchanged regions and thus obtain more accurate coarse change maps.

\noindent 
\textbf{Detection Results.} 
In addition to the visualizations of feature maps, we also provide the detected changed regions of our proposed VcT and other SOTA models. For better visualization, we use different colors to denote TP, TN, FP, and FN, i.e., white, black, red, and green color. To be specific, as shown in Fig.~\ref{visDetectionResults}, the (a), (b), and (c) column denotes the input image A, input image B and GT map, respectively. The (d)-(h) columns are the detected change results of other comparing methods, which are obviously worse than the proposed VcT (I). This fully demonstrates the advantages of our proposed VcT model for the remote sensing change detection.


\begin{figure}[t]
\centering
\includegraphics[width=0.5\textwidth]{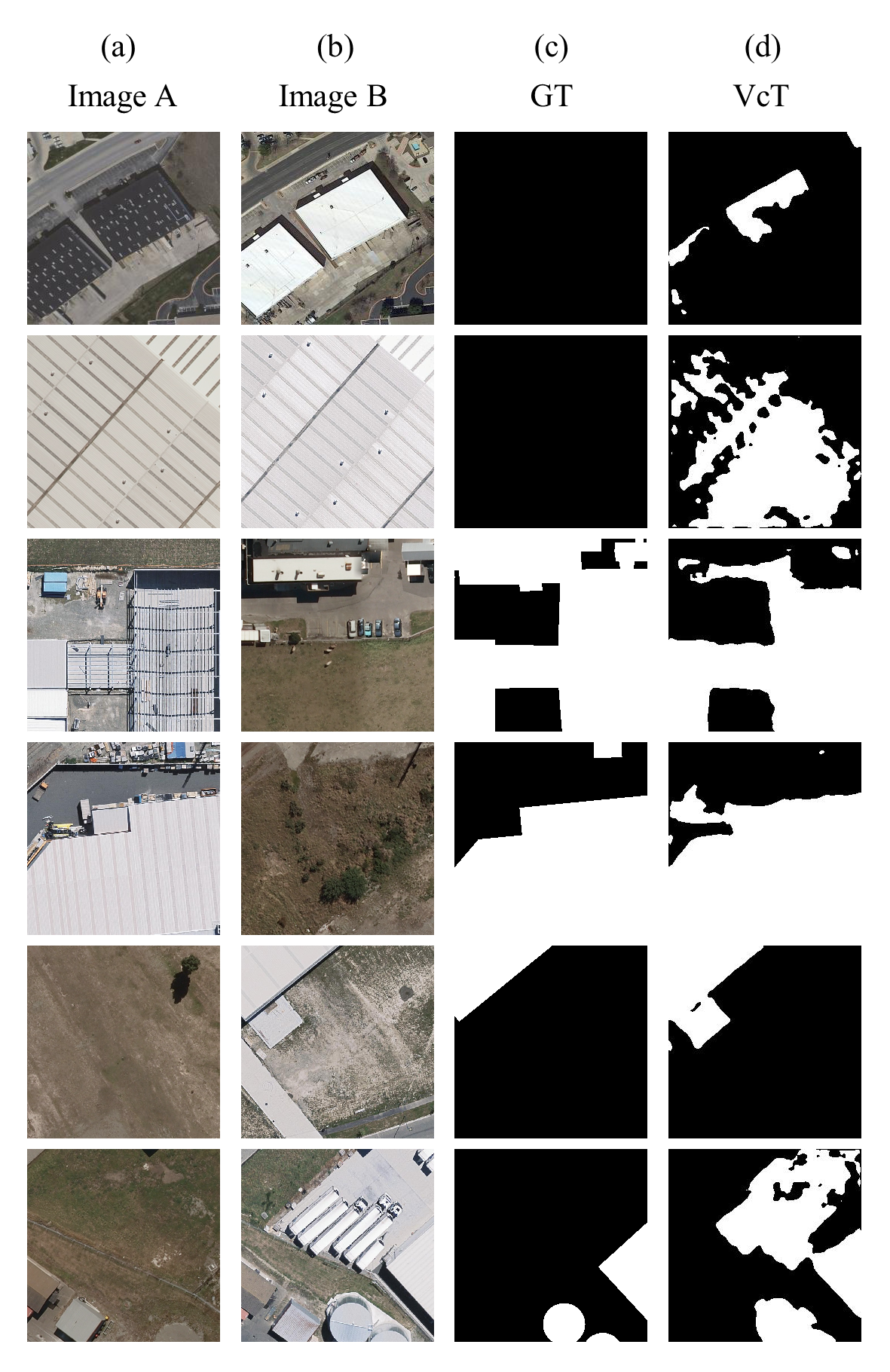} 
\caption{Limited detection results of our proposed VcT model.}
\label{limitation}
\end{figure}

\subsection{Limitation Analysis} \label{limitAnalysis}
Although our proposed VcT achieves good performance on existing remote sensing change detection datasets, however, it still can be improved from the following aspects. 
On the one hand, the top-K token selection in the RTM module works well in regular scenarios. When the changed regions are biased towards extreme cases, for example, there are too many changed regions or no changed regions at all, the fixed token selection strategy may bring us sub-optimal results only. Some failed cases can be found in Fig.~\ref{limitation} (1-4$^{th}$ row). These limited results may be addressed well if the number of selected tokens can be adaptively tuned.  
On the other hand, we find that many non-building regions (such as vehicles) are changed and our detector indeed finds these regions. But these datasets focus on detecting the changed buildings and ignore the others when annotating the ground truth labels. 
\textcolor{black}{Intuitively, the proposed method can observe higher detection accuracy if the complete changes are labeled, as shown in Fig.~\ref{limitation} (5-6-{th} row). }


In addition, for semantic information assistance, certain large-scale foundational models~\cite{wang2022empirical} can be utilized here. Examples include Grounding DINO \cite{liu2023grounding} and the Segment Anything Model (SAM) \cite{kirillov2023segment}. For example, in the case of a specific building detection dataset, text prompts can be employed to segment building regions by using pre-trained large-scale models. This approach allows the model to concentrate solely on detecting changes within the region of interest while disregarding irrelevant temporary changes in trees, vehicles, etc. Nevertheless, it is essential to acknowledge that there may be domain gaps between remote sensing images and natural images, potentially resulting in sub-optimal segmentation outcomes. Therefore, further experimental exploration is warranted in this regard. We leave them as our future works.

\section{Conclusion} \label{conclusion}
In this work, we propose a novel framework for remote sensing change detection, termed VcT. It mainly consists of three main modules, i.e., reliable token mining module,  Transformer module, and  prediction head. The backbone network is shared between two input images and to produce initial CNN features. Then, a coarse change map can be generated by considering a structured graph and top-K token selection, with diverse and accurate tokens mined via  K-means clustering in the coarse-to-fine manner. The Transformer layers are used to further enhance inter- and intra-relations between the tokens. Also, anchor-primary attention is adopted to achieve cross-fusion between enhanced and original features. Finally, a prediction head is adopted to transform the features into pixel-level change detection maps. We conduct extensive experiments on three datasets
to demonstrate the effectiveness and benefits of the proposed VcT.

\section*{Acknowledgement} 
This research is supported in part 
by Anhui Provincial Key Research and Development Program (2022i01020014); 
National Natural Science Foundation of China (62076004; 62102205); Natural Science Foundation of Anhui Province (2108085Y23).

{
\bibliographystyle{IEEEtran}
\bibliography{reference}
}

\end{document}